%% file: main.tex
\documentclass[10pt,twocolumn,letterpaper]{article}

\usepackage{cvpr}              
\usepackage{amssymb}
\usepackage{amsmath}
\usepackage{colortbl}
\usepackage{xcolor}
\definecolor{darkgreen}{RGB}{84,164,64} 
\definecolor{brightgreen}{RGB}{138,225,153}
\usepackage{makecell}
\usepackage{array}
\usepackage{graphicx}
\usepackage{caption}
\usepackage{enumitem}
\usepackage{pifont}
\usepackage{tikz}


\definecolor{cvprblue}{rgb}{0.21,0.49,0.74}
\usepackage[pagebackref,breaklinks,colorlinks,allcolors=cvprblue]{hyperref}

\def\paperID{14623} 
\def\confName{CVPR}
\def\confYear{2025}

\newcommand*{\MinNumber}{0.6}
\newcommand*{\MidNumber}{1.0} 
\newcommand*{\MaxNumber}{4}
\newcommand*{\Ratio}{100}
\usepackage{pgfplots}
\usepgfplotslibrary{dateplot}

\newcommand{\cl}[1]{%
        \ifdim #1 pt > \MidNumber pt
            \pgfmathsetmacro{\PercentColor}{max(min(\Ratio*(#1 - \MidNumber)/(\MaxNumber-\MidNumber),\Ratio),0.00)} %
            \edef\x{\noexpand\cellcolor{red!\PercentColor!yellow!100}}\x #1
        \else
            \pgfmathsetmacro{\PercentColor}{max(min(\Ratio*(\MidNumber - #1)/(\MidNumber-\MinNumber),\Ratio),0.00)} %
            \edef\x{\noexpand\cellcolor{green!\PercentColor!yellow!100}}\x #1
        \fi
}
\newcommand*{\MinN}{0.70}
\newcommand*{\MidN}{1.00} 
\newcommand*{\MaxN}{2.5}
\newcommand{\cla}[1]{%
        \ifdim #1 pt > \MidN pt
            \pgfmathsetmacro{\PercentColor}{max(min(\Ratio*(#1 - \MidN)/(\MaxN-\MidN),\Ratio),0.00)} %
            \edef\x{\noexpand\cellcolor{red!\PercentColor!yellow!80}}\x #1
        \else
            \pgfmathsetmacro{\PercentColor}{max(min(\Ratio*(\MidN- #1)/(\MidN-\MinN),\Ratio),0.00)} %
            \edef\x{\noexpand\cellcolor{green!\PercentColor!yellow!80}}\x #1
        \fi
}

\newcommand*{\MinB}{0.40}
\newcommand*{\MidB}{0.90} 
\newcommand*{\MaxB}{2.3}
\newcommand{\clb}[1]{%
        \ifdim #1 pt > \MidB pt
            \pgfmathsetmacro{\PercentColor}{max(min(\Ratio*(#1 - \MidB)/(\MaxB-\MidB),\Ratio),0.00)} %
            \edef\x{\noexpand\cellcolor{red!\PercentColor!yellow!80}}\x #1
        \else
            \pgfmathsetmacro{\PercentColor}{max(min(\Ratio*(\MidB- #1)/(\MidB-\MinB),\Ratio),0.00)} %
            \edef\x{\noexpand\cellcolor{green!\PercentColor!yellow!80}}\x #1
        \fi
}

\def\eg{\emph{e.g.}} 
\def\ie{\emph{i.e}} 
\newcommand{\Tref}[1]{Tab.~\ref{#1}}

\newcommand{\Fref}[1]{Fig.~\ref{#1}}
\newcommand{\Sref}[1]{Sec.~\ref{#1}}

\newcommand{\eref}[1]{Eq.~(\ref{#1})}
\newcommand{\fref}[1]{Fig.~\ref{#1}}
\newcommand{\sref}[1]{Sec.~\ref{#1}}

\definecolor{1}{RGB}{255, 174, 176}
\definecolor{2}{RGB}{255, 215, 172}
\definecolor{3}{RGB}{255, 255, 168}

\title{Spiking GS: Towards High-Accuracy and Low-Cost Surface Reconstruction via Spiking Neuron-based Gaussian Splatting}

\author{
Weixing~Zhang$^{1,2,}$\thanks{Co-first authors. ~~\ddag Corresponding author.}~
Zongrui~Li$^{3,4,*}$
De~Ma$^{1,5}$
Huajin Tang$^{1,5}$
Xudong Jiang$^{3,4}$
Qian Zheng$^{1,5,\ddag}$
Gang Pan$^{1,5}$
\smallskip 
\\
$^1${\small The State Key Lab of Brain-Machine Intelligence, Zhejiang University}\\
$^2${\small School of Software Technology, Zhejiang University}\\
$^3${\small Rapid-Rich Object Search (ROSE) Lab, Nanyang Technological University}\\
$^4${\small School of Electrical and Electronic Engineering, Nanyang Technological University}\\
$^5${\small College of Computer Science and Technology, Zhejiang University}\\
{\small \{weixingzhang,~made,~htang,~qianzheng,~gpan\}@zju.edu.cn},
{\small\{zongrui001,~EXDJiang\}@ntu.edu.sg}
}

\begin{document}

\input{sec/0_abstract}    
\input{sec/1_intro}
\input{sec/2_rela}
\input{sec/3_preli}
\input{sec/4_method}
\input{sec/5_exp}
\input{sec/6_conclusion}
{
    \small
    \bibliographystyle{ieeenat_fullname}
    \bibliography{main}
}

\input{sec/7_suppl}

\end{document}

%% file: sec/0_abstract.tex
\twocolumn[{%
	\renewcommand\twocolumn[1][]{#1}%
	\maketitle
    \centering
	\includegraphics[width=1\linewidth]{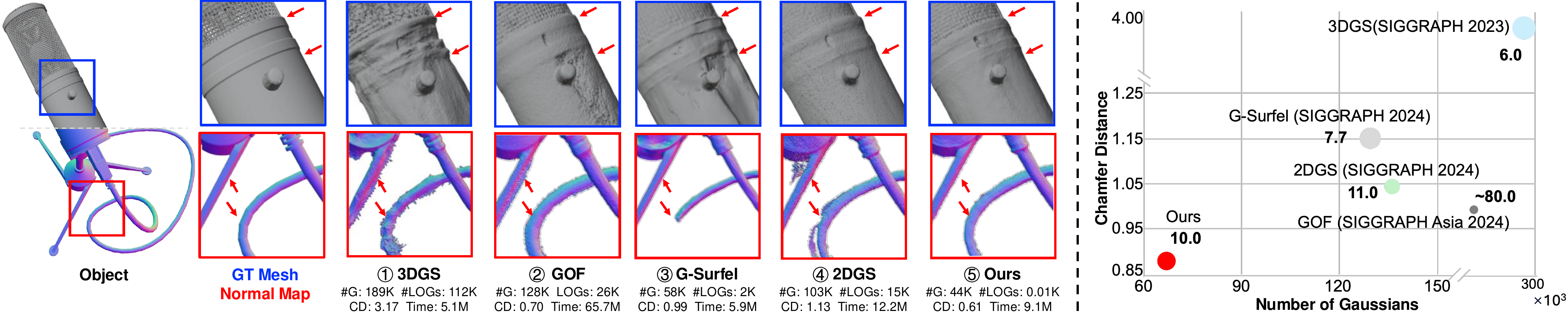}
    \captionof{figure}{\textbf{Left}: we show a visual comparison of the extracted mesh for the {\sc Mic} object from the NeRF-Synthetic dataset~\cite{mildenhall2021nerf}. We additionally compare the Gaussians' number (\#G), number of \textbf{L}ow-\textbf{O}pacity \textbf{G}aussians (\#\textbf{LOG}s), Chamfer Distance (CD), and optimization time between our and previous methods. We regard opacity lower than 0.2 as low-opacity for statistics. \textbf{Right}: we show statics of different methods on the NeRF-Synthetic dataset~\cite{mildenhall2021nerf} in terms of CD, \#G, and training time visualized by the size of the circles, bigger circles indicate faster speed; numbers indicate the specific time cost in minutes. }
    \label{fig:teaser}
}]

\begin{abstract}
3D Gaussian Splatting is capable of reconstructing 3D scenes in minutes. Despite recent advances in improving surface reconstruction accuracy, the reconstructed results still exhibit bias and suffer from inefficiency in storage and training. This paper provides a different observation on the cause of the inefficiency and the reconstruction bias, which is attributed to the integration of the low-opacity parts (LOPs) of 3D Gaussians. We identify the source of LOPs and propose Spiking GS to reduce two types of LOPs by integrating spiking neurons into the Gaussian Splatting pipeline. Specifically, we introduce global and local full-precision integrate-and-fire spiking neurons to the opacity and representation function of flattened 3D Gaussians, respectively. Furthermore, we enhance the density control strategy with spiking neurons' thresholds and a new cloning criterion. Our method can represent more accurate reconstructed surfaces at a lower cost. 
The code is available at \textcolor{cyan}{https://github.com/zju-bmi-lab/SpikingGS}.
\end{abstract}

%% file: sec/1_intro.tex
\section{Introduction}

3D surface reconstruction from multiview RGB images is a challenging task in computer graphics and vision~\cite{turkulainen2024dn,verbin2022ref}. Methods based on Neural Radiance Fields (NeRF)~\cite{mildenhall2021nerf,29guo2022nerfren,martin2021nerf,10regnerf,wang2021neus} can extract geometry information from an implicit representation but require extensive computational time for training, limiting their application scenarios. In contrast, 3D Gaussian Splatting (3DGS)~\cite{kerbl20233d} is exceptionally fast in 3D reconstruction, which is gaining more and more attention. 

In particular, 3DGS-based methods integrate the opacity of Gaussians for view rendering and surface reconstruction. In the optimal scenario~\cite{dai2024high,guedon2024sugar}, only a few {\it high-opacity} Gaussian tightly clustered around the surfaces contribute to integration.
However, we observe that a suboptimal situation often arises, where Gaussians with low- and high-opacity are distributed around the surfaces, yielding the emergence of excessive {\it \textbf{L}ow-\textbf{O}pacity \textbf{P}arts} (\textbf{LOP}s) in the 3DGS integration. Failing to suppress the emergence of LOPs results in excessive 3D Gaussians, hindering training and storage efficiency. 
Besides, LOPs also negatively influence surface reconstruction, leading to surface bias. Most of the previous 3DGS-based surface reconstruction methods overlook the effects of LOPs, focusing instead on improving the reconstructed surface's quality through extra regularization~\cite{dai2024high}, enhancing mesh extraction method~\cite{yu2024GOF}, and new Gaussian primitives~\cite{chen2023neusg,dai2024high,huang20242d}. Despite the effectiveness of these methods, there is still considerable potential to improve the accuracy and efficiency of reconstruction in the aspect of suppressing excessive LOPs. 

In this paper, we conduct a detailed analysis of the above observations in \Sref{sec:analyze-logs}, based on which we further identify two primary sources of LOPs: {\it \textbf{L}ow-\textbf{O}pacity \textbf{G}aussians} (\textbf{LOG}s) and the {\it \textbf{L}ow-\textbf{O}pacity \textbf{T}ails} (\textbf{LOT}s) of Gaussians (intuitive interpretation shown in \fref{fig:lops1}). To handle LOGs and LOTs, we implement a global full-precision integrate-and-fire (FIF) spiking neuron on the opacities of Gaussians and a local FIF spiking neuron considering the Gaussian representation itself, respectively. The former uses a learnable threshold to raise the lower bound of the opacity value of the scene, which effectively reduces the number of LOGs; the latter introduces the discontinuity in the Gaussian representation function, which reduces the contribution of LOTs in 3DGS integration. By suppressing the integration of LOPs for view rendering, we can adaptively regulate LOPs, facilitating robust convergence towards the optimal scenario for various scenes. 
Although Spiking GS excels in surface reconstruction, it may overly remove Gaussians in occluded regions, which slightly harms the reconstruction accuracy. To avoid this, we introduce the scale-based clone to compensate for the missing Gaussians in these regions. Moreover, we implement several regularization losses to further enhance the method's performance. 
Extensive experiments show that Spiking GS achieves comparable or superior 3D reconstruction performance on mainstream datasets with much fewer Gaussians and significantly outperforms other methods in semi-transparent scenarios. A glimpse of Spiking GS's performance is shown in~\fref{fig:teaser}. 
In summary, we make the following contributions:
\begin{itemize}
     \item We analyze the formation of LOPs, which consisting of LOGs and LOTs, are prevalent in the generated results of previous methods. We further show that the integration of excessive LOPs bring the bias of surface reconstruction and the high-cost of the optimiztion.
     \item We implement a global FIF spiking neuron on the opacity of Gaussians and a local FIF spiking neuron on each Gaussian representation function, which reduces LOPs.
     \item We propose a scale-based clone method and several regularization losses to improve reconstruction accuracy. The proposed Spiking GS achieves higher reconstruction accuracy with fewer Gaussians for low storage occupation and high optimization efficiency.
\end{itemize}

%% file: sec/2_rela.tex
\section{Related Work}
\textbf{Surface Reconstruction with Implicit Representation.} Neural Radiance Field~\cite{mildenhall2021nerf} has explored a novel way to reconstruct 3D surface based on a neural radiance field. However, such reconstruction suffers from significant artifacts due to insufficient geometric constraints~\cite{verbin2022ref,wang2021neus,yariv2021volume,fu2022geo,wang2022hf,wang2023pet} and defects of MLP in representing real-world scene~\cite{liao2024spiking,reiser2024binary}. To improve the reconstruction quality, some of the subsequent works focus on modifying the MLP-based representation by either integrating with Signed Distance Functions (SDFs), a more explicit surface representation~\cite{wang2021neus,yariv2021volume,fu2022geo,wang2022hf,wang2023pet}, or introducing discontinuities into the density field~\cite{liao2024spiking,reiser2024binary}. Other works adopt strong priors to regularize the reconstruction, such as priors from depth~\cite{deng2022depth,yu2022monosdf} and MVS~\cite{zhang2021learning,zuo2022view}. Nevertheless, these methods still require prolonged time for optimizing the neural radiance fields. In contrast, our method is developed on the fast 3DGS method to achieve more accurate surface reconstruction with lower cost.

\noindent \textbf{Surface Reconstruction with 3D Gaussians.} Although 3D Gaussian Splatting~\cite{kerbl20233d} achieves high fidelity on novel view synthesis, it struggles to reconstruct accurate surface due to the difficulty of aligning primitives with the actual surface~\cite{dai2024high} and the inconsistency between mesh extraction and view rendering~\cite{yu2024GOF}. To address these issues, ~\cite{chen2024vcr,guedon2024sugar} utilize a regularization to flatten the 3D Gaussian ellipsoids into flat shapes, while \cite{dai2024high,huang20242d} directly use flattened 3D Gaussians as primitives. Others redesigned the opacity modeling of Gaussians~\cite{lyu20243dgsr,yu2024GOF}, improving the details of the reconstructed surfaces. However, these approaches still face challenges in the accuracy of surface reconstruction in scenes including objects with complex materials (\ie, semi-transparent). 
The proposed Spiking GS identifies a widely overlooked correlation between the reconstruction bias and LOPs, achieving considerable improvement on the reconstruction quality and efficiency by reducing LOPs.

\noindent \textbf{Efficient Geometry Representation of 3D Gaussians.} 
The use of a large number of Gaussians to accurately represent scenes often results in inefficient training and storage. To address this, recent efforts have focused on utilizing compact feature encodings and pruning strategies to improve efficiency in Gaussian representations~\cite{chen2024hac,lee2024compact,lu2024scaffold}. However, these methods compromise surface reconstruction accuracy. Other works~\cite{dai2024high,guedon2024sugar} eliminate the occurrence of LOGs or introduce opacity-based regularization terms to reduce the extra Gaussians. However, they suffer from limited representation capabilities, particularly in scenes with semi-transparent objects, or limited effect on the occurrence of Gaussians. 
Although our method shares a similar target of reducing the number of Gaussians for improved efficiency, we specifically focus on minimizing the number of LOPs in integration. As a result, we not only effectively achieve the target but also maintain the representation capabilities for accurate surface reconstruction in various scenes.

\noindent \textbf{Spiking Neurons in Surface Reconstruction.} Spiking Neural Networks (SNNs) have received extensive attention recently in many fields due to their energy efficiency~\cite{kim2022privatesnn,ren2024spiking}. It has been utilized in processing video or event signals~\cite{paredes2021back,zhu2022event}, classification tasks~\cite{li2022brain,zhang2022spiking}, and adversarial attacks~\cite{ding2022snn,kundu2021hire,liang2022toward,zhou2021temporal}. Recently,~\citet{liao2024spiking} introduce the spiking neuron into the NeRF-based neural surface reconstruction for better performance in representing discontinuities of the scene. 
The proposed spiking GS is also equipped with the spiking neurons but for its efficiency in removing the LOPs in 3DGS. Experiments show that it significantly improves the performance of 3DGS-based surface reconstruction.

%% file: sec/3_preli.tex
\begin{figure}[t]
    \centering
    \includegraphics[width=0.48\textwidth]{"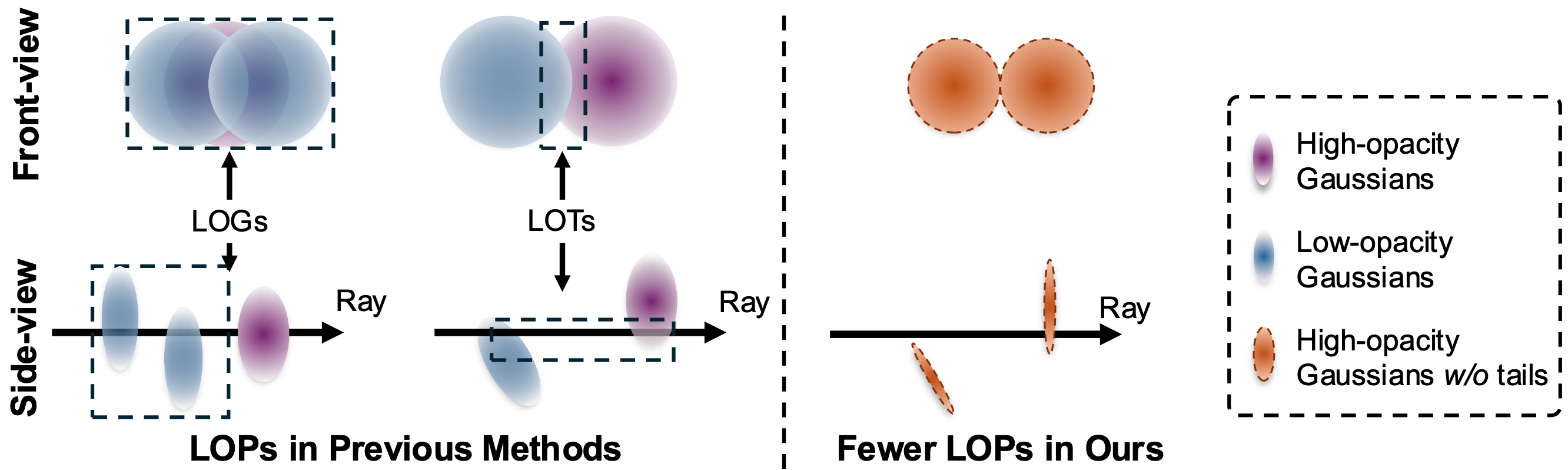"} 
    \caption{\textbf{Illustration of LOPs (LOGs and LOTs)}. From top to bottom, the first (second) row shows the front (side) view of the Gaussians around the surface.}
    \label{fig:lops1}
\end{figure}

\section{Preliminaries and Analysis}
In this section, we present preliminaries in Spiking GS, including flattened Gaussian splatting and spiking neurons. We then conduct an analysis of LOPs.

\subsection{Preliminary}
\label{sec:preliminary}

\textbf{Flattened Gaussian Splatting}.
3DGS~\cite{kerbl20233d} use a set of 3D Gaussian primitives $\mathcal{G}$ to model the real world scene, each of which is parameterized by an opacity $\alpha \in[0,1]$, central point's position $\mathbf{p} \in \mathbb{R}^{3 \times 1}$, scale vector $\mathbf{S} \in \mathbb{R}^{3 \times 1}$, and the rotation matrix $\mathbf{R} \in \mathbb{R}^{3 \times 3}$ parameterized by a quaternion $\mathbf{r}$. In this paper, we apply flattened 3D Gaussians~\cite{dai2024high} as the 3D primitive due to its ability to reconstruct richer surface details than other methods. 
The distribution of flattened 3D Gaussian is defined by:
\begin{equation}
    \label{eq:1}
    \mathcal{G}(\mathbf{x})=e^{-\frac{1}{2}\left(\mathbf{x}-\mathbf{p}\right)^\top \Sigma^{-1}\left(\mathbf{x}-\mathbf{p}\right)},
\end{equation}where $\mathbf{x}$ is the 3D coordinate, $\Sigma$ is the covariance matrix expressed as $\mathbf{R}\mathbf{S}\mathbf{S}^\top\mathbf{R}^\top$, set as  $[\mathbf{S}^\mathrm{x}, \mathbf{S}^\mathrm{y}, 0]^\top$~\cite{dai2024high}.
In the rendering process, Gaussians in the world coordinates are transformed into the camera coordinates using a world-to-camera transformation matrix $\mathbf{W}$ and projected onto the image plane using an affine approximation matrix $\mathbf{J}$~\cite{zwicker2001surface}:
\begin{equation}
    \label{eq:2}    \Sigma^{\prime}=\mathbf{J}\mathbf{W}\Sigma\mathbf{W}^\top\mathbf{J}^\top,
\end{equation}
where the third row and column of $\Sigma$ is skipped for projection. After that, the rendered color of each pixel is calculated via alpha-blending of $N$ ordered Gaussians overlapped on pixel points $\mathbf{u}$:
\begin{equation}
   \label{eq:alpha-blending}
   \mathbf{c}(\mathbf{u})=\sum_{i=0}^N T_i \omega_i \mathbf{c}_i, \quad T_i=\prod_{j=0}^{i-1}\left(1-\omega_j\right), \quad \omega_i=\mathcal{G}_i^{\prime} \alpha_i,
\end{equation}
where $\mathbf{c}(\mathbf{u})$ is the pixel value at point $\mathbf{u}$, $\mathbf{c}_i$ is the RGB color of Gaussians, $\mathcal{G}_i^{\prime}\left(\mathbf{u}\right)=e^{-\frac{1}{2}(\mathbf{u}-\mathbf{p})^{\top} \Sigma^{\prime-1}(\mathbf{u}-\mathbf{p})}$ is the Gaussians projected in 2D, $\alpha_i$ is the Gaussian opacity, $\omega_i$ is the weighted Gaussian opacity. During training, Gaussians' attributes are optimized by a photometric loss, and the number of Gaussians is controlled by an adaptive density control process~\cite{kerbl20233d}. 

\begin{figure}[t]
    \centering
    \includegraphics[width=0.48\textwidth]{"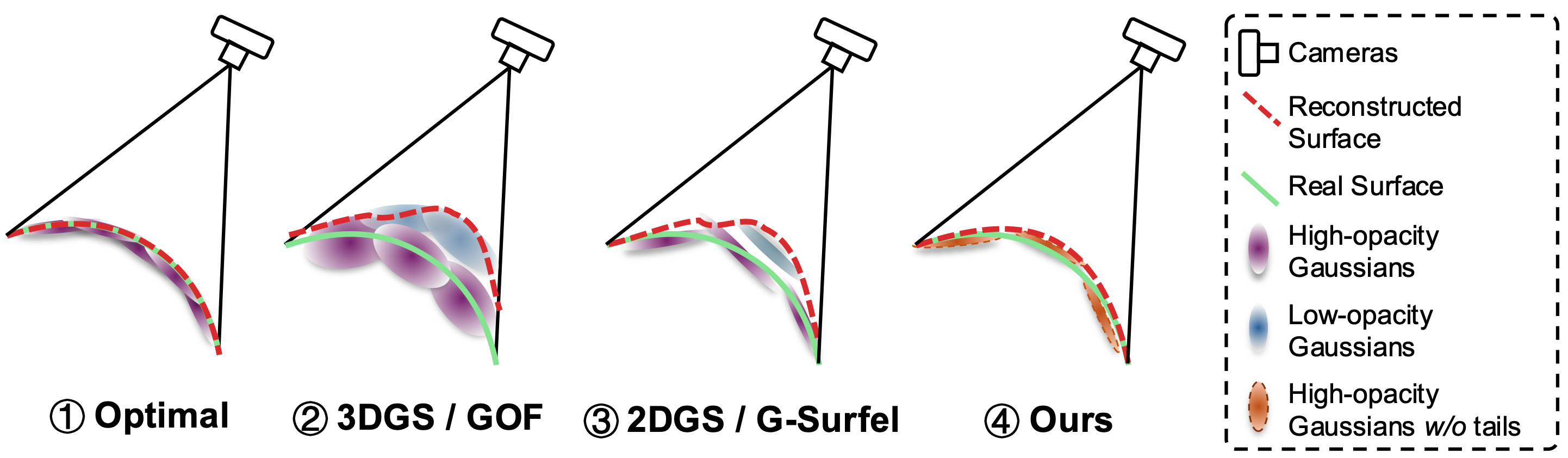"} 
    \caption{\textbf{Visualization of the side view of the optimized Gaussian around the surface of interest under different situations}: \textcircled{\raisebox{-0.9pt}{1}} optimal (according to~\cite{dai2024high,guedon2024sugar}), \textcircled{\raisebox{-0.9pt}{2}}\textcircled{\raisebox{-0.9pt}{3}} suboptimal, and \textcircled{\raisebox{-0.9pt}{4}} ours. The bias of the reconstructed surface can be observed by the distance between the \textcolor{red}{red} and \textcolor{brightgreen}{green} lines.}
    \label{fig:lops2}
\end{figure}

\begin{figure*}[t]
    \centering
    \includegraphics[width=1.0\textwidth]{"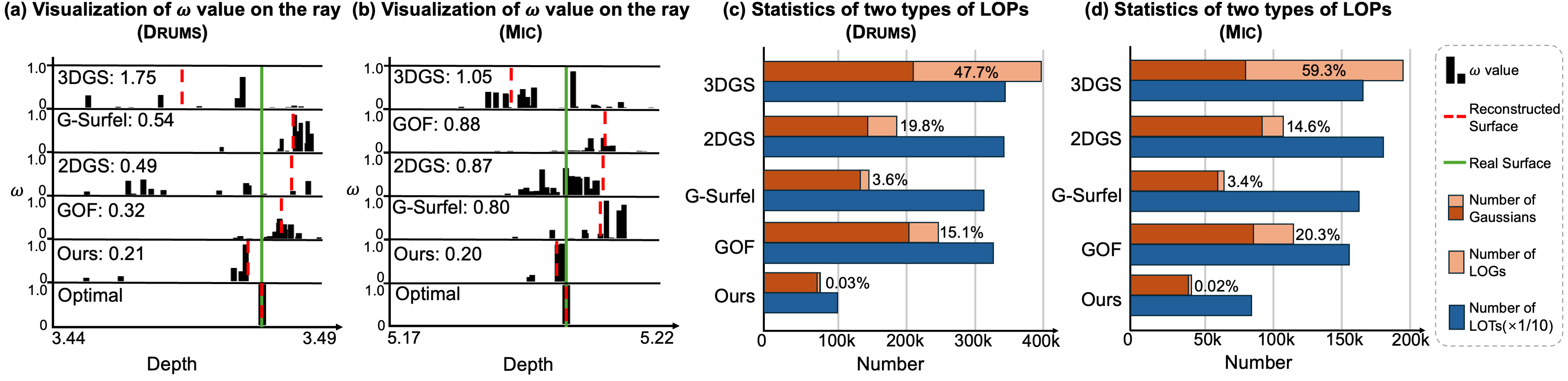"} 
    \caption{\textbf{Visualization of $\omega$ value on a view ray of (a) {\sc Drums} and (b) {\sc Mic} from NeRF-Synthetic Dataset}~\cite{mildenhall2021nerf}. We select representative patterns of $\omega$ near the actual surface. The height of each bin represents the $\omega$ value, the \textcolor{red}{red} and \textcolor{darkgreen}{green} lines represent the GT depth and the estimated depth. Short bins indicate the presence of LOPs spreading out around the actual surface positions. Numbers indicate the corresponding depth errors. \textbf{Statistics on two types of LOPs on (c) {\sc Drums}, (d) {\sc Mic} from NeRF-Synthetic Dataset}, including the number of Gaussians, the proportion of LOGs, and the number of LOTs (averaged overlaps of Gaussian’s tails with a Gaussian representation function value below 0.1 in multiple views).}
    \label{fig:bar_chart}
\end{figure*}

\noindent \textbf{Spiking Neuron}.
We implement a simplified full-precision\footnote{Following previous work, we note the spiking neuron with float rather than binary output as ``full-precision'' spiking neuron.} integrate-and-fire (FIF) spiking neuron~\cite{li2022brain, liao2024spiking}. The model is described as:
\begin{equation}
     o = l s(l),~~~s =
    \begin{cases}
        0&l<\bar{V}, \\
        1&\text{otherwise},
    \end{cases}
\end{equation}
where $l$ is the input value, $\bar{V}$ is a spiking threshold, $o$ is the output value. To make the spiking neuron differentiable, we use the surrogate gradient~\cite{li2021differentiable,liao2024spiking} to calculate the gradient during training as:
\begin{align}
    \begin{aligned}
    \frac{\partial o}{\partial l} = s, ~~{\frac{\partial o}{\partial \bar{V}}} = \lambda l \max(0,\frac{k-\left|l - \bar{V}\right|} {k^2}), \\
    \label{eq:surrogate}
    \end{aligned}
\end{align}
where, $k$ and $\lambda$ are hyperparameters for surrogate gradients.

\subsection{Analyzing Low-opacity Parts}
\label{sec:analyze-logs}
Upon analyzing the pipelines and generated Gaussians of 3DGS methods, we obtain insights below.

\noindent\textbf{The suboptimal scenario leads to the emergence of excessive LOPs.} As shown in \eref{eq:alpha-blending}, the pixel value in a given view is computed by integrating the color and opacity of the nearby projected Gaussians. If we focus on reducing the loss between the rendered and ground truth pixel values $\mathbf{c}(\mathbf{u})$ while overlooking regularizing the contribution of each Gaussian, there exist two possible scenarios for deriving the same pixel value $\mathbf{c}(\mathbf{u})$: the scenario where only a few high-opacity Gaussians are involved, or the scenario where both a few high-opacity Gaussians and excessive low-opacity Gaussians are involved. According to~\cite{dai2024high,guedon2024sugar}, the latter is suboptimal. We observe that existing state-of-the-art (SOTA) methods tend to converge to such suboptimal scenarios, as illustrated in \Fref{fig:lops2}. We validate this observation by visualizing the representative pattern of the $\omega$ values along the view rays generated by SOTA methods in \Fref{fig:bar_chart} (a) and (b), showing that excessive LOPs are introduced in the SOTA methods. This is further validated by the statistic results in \Fref{fig:bar_chart} (c) and (d).

\noindent\textbf{Excessive LOPs bring the bias of surface reconstruction} as extracting surfaces from Gaussians involves similar opacity integration in \eref{eq:alpha-blending}, as shown in \Fref{fig:lops2}.
Take the widely used TSDF fusion~\cite{huang20242d} as an example. It first computes the depth for every pixel point $\mathbf{u}$ as:
\begin{equation}
   \label{eq:alpha-blending-d}
    t(\mathbf{u})=\sum_{i=0}^N \prod_{j=0}^{i-1}\left(1-\mathcal{G}_j^{\prime} \alpha_j\right) \mathcal{G}_i^{\prime} \alpha_i t_i,
\end{equation}
where $t_i$ is the depth of the $i$-th 3D Gaussians. According to \eref{eq:alpha-blending-d}, the computed depth may be biased by redundant LOPs. We visualize the estimated bias along the ray in \Fref{fig:bar_chart} (a) and (b).
The error in the depth map will be passed to the extracted mesh since the TSDF fusion method fuses depth maps from different views to generate the final mesh. 
Such an issue also exists in other advanced surface extraction methods, \eg, the tetrahedral-based surface extraction proposed in~\cite{yu2024GOF}, due to their similar computation method in \eref{eq:alpha-blending-d} for the opacity field derivation. Although we have shown that numerous LOPs bring the bias of surface reconstruction, we would like to emphasize that they are not the only cause of the bias. Due to a training process driven by photometric loss instead of direct regularization on geometry, it is possible that the reconstructed surfaces still exhibit high reconstruction bias even with fewer LOPs. For example, G-Surfel sometimes shows a high depth error but with fewer LOPs compared to other methods (\fref{fig:bar_chart} (a)).

\noindent \textbf{LOPs consist of LOGs and LOTs.} As $\omega=\mathcal{G}^{\prime} \alpha$ (\eref{eq:alpha-blending}), either a low $\alpha$ or a low $\mathcal{G}^\prime$ result in a low $\omega$ value, which forms the LOPs in the 3DGS integration. Note that there will be an overlap between these two types of LOPs, as LOTs exist in every Gaussian due to the continuity of Gaussian representation function. 
Statistics results confirm that both LOGs and LOTs contribute to the 3DGS integration in previous methods (\fref{fig:bar_chart} (c) (d)), leading to bias and inefficiency during the optimization. By utilizing spiking neurons, our method effectively reduces the number of LOPs, significantly improving the accuracy and efficiency in surface reconstruction.

%% file: sec/4_method.tex
\section{Method}
    \label{sec:4}
\begin{figure}[t]
    \centering
    \includegraphics[width=0.47\textwidth]{"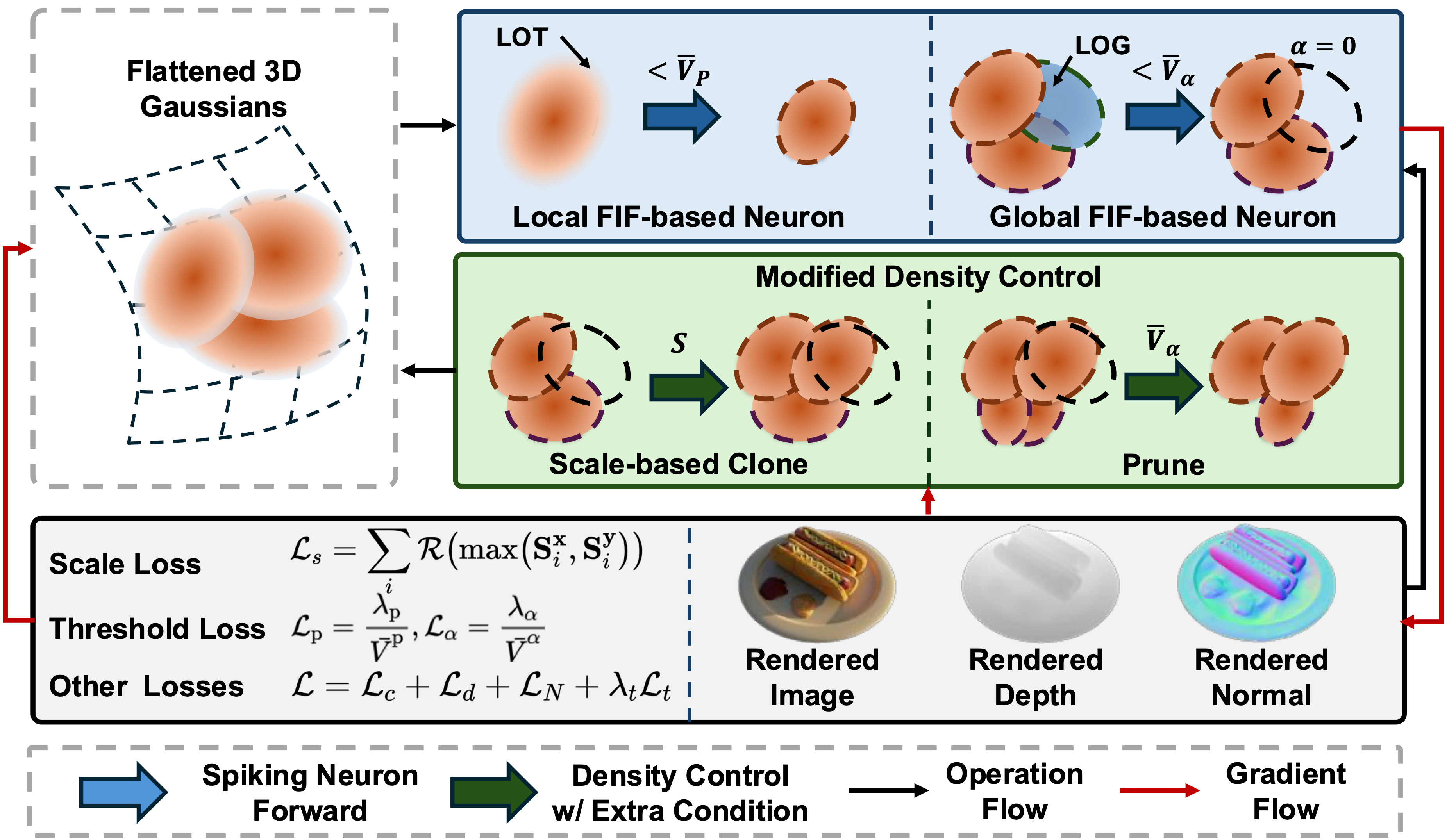"} 
    \caption{\textbf{The pipeline of the proposed Spiking GS.} Two types of FIF neurons are applied to the flattened Gaussians. A modified density control process can more adaptively adjust the number of Gaussians. Attributes of the Gaussians are updated by photometric loss $\mathcal{L}_{c}$, total variance loss $\mathcal{L}_{t}$, depth distortion loss $\mathcal{L}_{d}$, normal loss $\mathcal{L}_{N}$, scale loss $\mathcal{L}_{s}$, and threshold loss $\mathcal{L}_\alpha$, $\mathcal{L}_\mathrm{p}$. 
    }
    \label{fig:pipeline}
\end{figure}

This section presents details of the proposed Spiking GS. We first illustrate how the proposed global and local FIF neurons help reduce the number of LOPs (\Sref{sec:4.1}). In \Sref{sec:4.2}, we present the training process of Spiking GS to achieve high-accuracy surface reconstruction with a small number of Gaussians. The pipeline is shown in \fref{fig:pipeline}.

\subsection{FIF Neurons in Spiking GS}
\label{sec:4.1}
The key of the proposed Spiking GS method is leveraging FIF neurons to reduce the number of LOPs in integration for view rendering and surface reconstruction. Specifically, we employ the global FIF neuron to manage LOGs, and the local FIF neuron to control LOTs. These spiking neurons guide the training process, encouraging Gaussians to converge towards the optimal scenario, where fewer Gaussians, with minimal overlap in their tails, represent the scene (\fref{fig:teaser} (a) \textcircled{\raisebox{-0.9pt}{4}}, \fref{fig:teaser} (b) right). This brings improved accuracy and efficiency in surface reconstruction.

\noindent \textbf{Global FIF neuron on opacity}.  The global FIF neuron is implemented on the Gaussians' opacity $\alpha$. The output of FIF neuron is calculated as:
\begin{equation}
    \label{eq:8}
    \hat{\alpha}=
    \begin{cases}
        0&\alpha<\bar{V}^{\alpha} \\
        \alpha&\text{otherwise},
    \end{cases}
\end{equation}
where $\bar{V}^{\alpha}$ is shared by all Gaussians and optimized during training. Previous 3DGS methods~\cite{kerbl20233d, huang20242d} control the density of Gaussians by pruning and opacity resetting. Specifically, Gaussians with opacity values below a threshold $\epsilon_\alpha$ are removed during optimization, and their opacities are periodically reset to a low value, $\epsilon^\mathrm{r}_\alpha$. Although this density control can reduce LOGs near the camera, it is less effective at managing LOGs around surfaces. In contrast, the proposed global FIF neuron leverages the discontinuity and differentiability of spiking neurons equipped with surrogate gradient to provide a more adaptive control over LOGs during training. Specifically, Gaussians are set to zero opacity when their values fall below a learnable threshold $\bar{V}^\alpha$. These Gaussians, often corresponding to LOGs, are then removed in pruning. Additionally, we replace $\epsilon^\mathrm{r}_\alpha$ with $\bar{V}^{\alpha}$ in opacity-resetting, where $\bar{V}^{\alpha}$ is consistently greater than $\epsilon^\mathrm{r}_\alpha$. This higher opacity-resetting threshold increases the overall opacity in scenes, which implicitly guides 3DGS to generate less LOGs.

\noindent \textbf{Local FIF neuron on representation function}.
The tails of Gaussians (low function value parts, \ie, LOTs) often overlap when they are positioned closely together. This overlap can sometimes contribute to the integration for view rendering and surface reconstruction, resulting in biases and an excessive number of Gaussians. To eliminate the contribution of LOTs, we use the local FIF neuron to set the value of the low-function-value parts to zero. Specifically, the Gaussian's representation function is defined as:

\begin{equation}
    \label{eq:9}
    \hat{\mathcal{G}}_i(\mathbf{x})=
    \begin{cases}
        0&\mathcal{G}_i(\mathbf{x})<\bar{V}^{\mathrm{p}}_i \\
        e^{-\frac{1}{2}\left(\mathbf{x}-\mathbf{p}\right)^\top \Sigma^{-1}\left(\mathbf{x}-\mathbf{p}\right)}&\text{otherwise},
    \end{cases}
\end{equation}where $\bar{V}^{\mathrm{p}}$ is the cut-off threshold set as a learnable parameter. Notably, each Gaussian is assigned a unique, learnable $\bar{V}^{\mathrm{p}}_i$, reflecting the fact that different regions in the scene may need varying cut-off thresholds to capture the geometry accurately. For instance, areas near object boundaries might necessitate a larger cut-off threshold to model the discontinuity at the boundary than other parts of the object. In practice, to better integrate $\hat{\mathcal{G}}$ into the rasterization framework, we directly apply the local FIF neuron on the projected 2D Gaussian on the screen, making it act {\it as if} applying the model on the 3D Gaussian. 

\subsection{Optimizing Spiking GS}
\label{sec:4.2}
Despite the success in using fewer Gaussians to represent the scene, artifacts may appear on the reconstructed surfaces, as a tighter clustering of Gaussians around the surfaces is required. To address this, additional regularization on the reconstructed surfaces is necessary. Furthermore, we observe that the proposed FIF neurons and the enhanced Gaussian pruning strategy may excessively remove Gaussians in regions with limited constraints (\eg, blind spots or areas captured by only a few views), which negatively impacts reconstruction accuracy. To mitigate this, we introduce a new criterion in the original cloning process, we call it the scale-based clone. The following section summarizes the loss functions implemented and explains the scale-based cloning process.

\noindent\textbf{Threshold loss.} This is proposed to enhance the effect of local and global FIF neurons. Two regularization terms $\mathcal{L}_\alpha$ and $\mathcal{L}_\mathrm{p}$ is introduced to increase $\bar{V}^{\alpha}$ and $\bar{V}^{\mathrm{p}}$ during training:
\begin{equation}
    \label{eq:16}
    \mathcal{L}_\alpha=\lambda_\alpha\frac{1}{\bar{V}^{\alpha}},~~\mathcal{L}_\mathrm{p}=\lambda_\mathrm{p}\frac{1}{\bar{V}^\mathrm{p}},
\end{equation}
where $\lambda_\alpha=\lambda_\mathrm{p}=2e^{-5}$. We experimentally find this hyperparameter setup fit general cases. Large values will cause Gaussians to be overly removed. Overly small value weakens the influence of spiking neurons.

\noindent\textbf{Scale loss.} As presented in~\cite{fan2024trim}, large-sized Gaussians tend to magnify minor errors during training and hinder the Gaussians' ability to capture high-frequency geometric details. Motivated by this insight, we propose a simple scale loss to constrain the scale of Gaussians:
\begin{equation}
    \label{eq:17}
    \mathcal{L}_s=\sum_{i} \mathcal{R}\left(\text{max}(\mathbf{S}^\mathrm{x}_i, \mathbf{S}^\mathrm{y}_i)\right), \mathcal{R}(\mathrm{m})= \begin{cases}0 & \mathrm{m}<\bar{V}^{\theta} \\ \mathrm{m} & \text { otherwise,}\end{cases}
\end{equation}
where $\bar{V}^{\theta}$ is a hyperparameter that controls the extent of scale regularization.

\noindent \textbf{Total loss.} Excluding the losses on $\bar{V}^{\alpha}$, $\bar{V}^{\mathrm{p}}$, and $\mathbf{S}$, we implement losses of previous methods. Except for the photometric loss $\mathcal{L}_{c}$ following~\cite{kerbl20233d}, we also use depth distortion loss $\mathcal{L}_{d}$, normal loss $\mathcal{L}_{N}$ following~\cite{huang20242d}, and total variance loss $\mathcal{L}_{t}$ following~\cite{karnieli2022deepshadow} to regularize the geometry. To sum up, the total loss $\mathcal{L}$ is calculated as:
\begin{equation}
    \label{eq:12}
    \mathcal{L}=\mathcal{L}_{c}+\mathcal{L}_{d}+\mathcal{L}_{N}+\lambda_{\alpha}\mathcal{L}_{\alpha}+\lambda_{\mathrm{p}}\mathcal{L}_{\mathrm{p}}+\lambda_s\mathcal{L}_{s}+\lambda_t\mathcal{L}_{t},
\end{equation}
where $\lambda_s$ and $\lambda_t$ are the hyperparameters set as $5e^{-4}$ and $1.0$ respectively.

\noindent \textbf{Scale-based Clone}. The geometry of blind spots or regions observed by limited viewpoints is particularly difficult to reconstruct due to the lack of constraints from viewpoints. Additionally, we find Gaussians in those regions hardly meet the clone criteria of splitting and cloning, which may be overly removed by the enhanced pruning strategy, resulting in deficiency (\eg, undesired holes and pits) in the reconstructed surface. To mitigate this issue, we propose a scale-based clone. Specifically, it additionally clone Gaussians that meet $\text{max}(\mathbf{S}^\mathrm{x}_i, \mathbf{S}^\mathrm{y}_i) \in[\bar{V}^\theta-\delta,\bar{V}^\theta+\delta]$, where $\delta=\frac{\bar{V}^\theta}{200}$. The rationale behind this approach is that blind-spot regions lack view constraints, resulting in small gradients in view-space positions. Consequently, Gaussians in these regions are often not cloned appropriately. We select $\bar{V}^\theta$ (the scale loss's hyperparameter) since the scale loss is view-invariant and becomes the dominant term in these regions. Thus, Gaussians with scales near this value are likely to be located in blind-spot regions. The new clone rule enriches the number of Gaussians in these regions, and mitigates artifacts in reconstructed such as holes.

%% file: sec/5_exp.tex
\section{Experiments}
\begin{figure*}[t]
    \centering
    \includegraphics[width=1\textwidth]{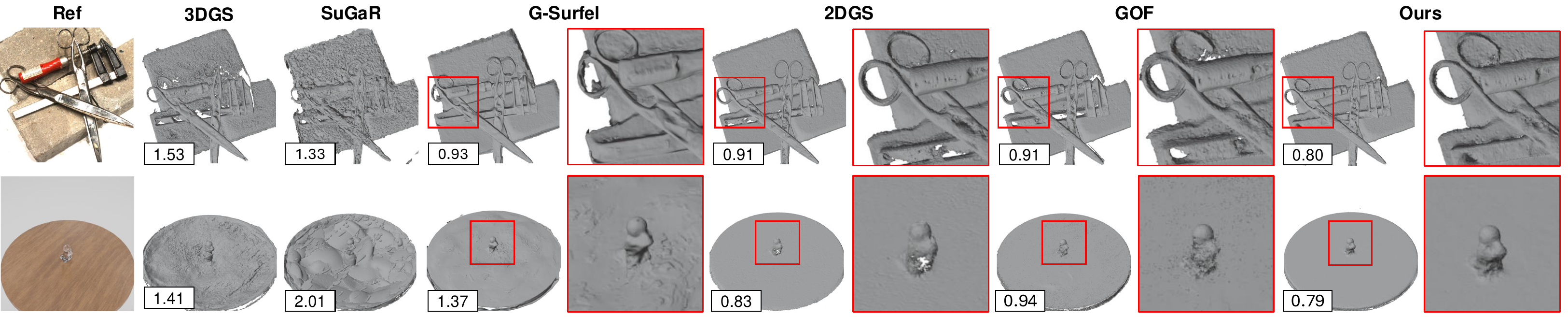}
    \caption{\textbf{Qualitative Comparisons of surface reconstruction} performed on DTU~\cite{jensen2014large} and Dex-NeRF datasets \cite{ichnowskidex}. We show the Chamfer distance in the bottom left corner of the image. 
}
\label{fig:mainexp}
\end{figure*}

Our training framework is developed on 3DGS~\cite{kerbl20233d} with flattened Gaussians~\cite{dai2024high} as primitive\footnote{For parameters setup and training details, please refer to the appendix.}.

\noindent\textbf{Evaluation metrics and datasets.} We test our method in 8 scenes from NeRF-Synthetic dataset~\cite{mildenhall2021nerf}, 6 scenes from Dex-NeRF dataset~\cite{ichnowskidex}, and 15 scenes from DTU dataset~\cite{jensen2014large}. These datasets include both synthetic and real-world scenes. To evaluate the accuracy of the 3D surface reconstruction, we use the Chamfer distance as the metric~\cite{dai2024high,huang20242d,liao2024spiking}. To evaluate the rendering quality, we use the standard PSNR, SSIM, and LPIPS metrics and report the results on NeRF-Synthetic and Dex-NeRF datasets on the test set. We also report training time and Gaussians' number (noted as \#G) to evaluate the efficiency of our method.

\noindent\textbf{Baselines.} We compare our method's performance with one NeRF-based neural surface reconstruction method (Spiking NeRF~\cite{liao2024spiking}, noted as ``SpNeRF'') and five Gaussian-based surface reconstruction methods (\ie, 3DGS~\cite{kerbl20233d}, SuGaR~\cite{guedon2024sugar}, G-Surfel~\cite{dai2024high}, 2DGS~\cite{huang20242d}, and GOF~\cite{yu2024GOF}). To further evaluate the training and storage efficiency, we compare our method with two compression methods (\ie, Compact-3DGS~\cite{lee2024compact}, denoted as ``C3DGS'', and HAC~\cite{chen2024hac}).

\subsection{Geometry Reconstruction}
\textbf{Quantitative Comparison.} To quantitatively assess the quality of the reconstructed surface, we compared the proposed Spiking GS with other methods on the NeRF-Synthetic~\cite{mildenhall2021nerf}, Dex-NeRF~\cite{ichnowskidex}, and DTU~\cite{jensen2014large} datasets. The results are shown in Tab. \ref{tab:1}-\ref{tab:3}. 
respectively. According to the results, our method achieves SOTA performance on most datasets (NeRF-Synthetic and Dex-NeRF). We find that the adaptability of the FIF neurons improves the method's robustness, showing stability in various scenes, especially with semi-transparent objects. However, we also notice that the advantage diminishes on the DTU dataset primarily because the viewpoints in the DTU are concentrated on the front side of the object, with few views from the back. This results in limited constraints introduced from the back views, negatively impacting our method.

\noindent\textbf{Qualitative Comparison.} To evaluate the quality of the reconstructed surfaces, we present the results of all methods for selected objects from two datasets in~\fref{fig:mainexp}.  While methods like 3DGS and SuGaR exhibit some high-frequency noise on the reconstructed surfaces, our method achieves both smoothness and preservation of intricate surface details. The advantage is obvious on the surface with complex materials (\eg, the reflective surface of the metallic object and the glass object in~\fref{fig:mainexp}). We attribute the improvements to the proposed spiking neurons, which adaptively guide Gaussians toward convergence in the optimal scenario.

\begin{table*}[t]
  \centering
  \caption{Quantitative Comparison on NeRF-Synthetic dataset~\cite{mildenhall2021nerf}. We report the Chamfer distance ($\times\ 10^{-2}$) for the reconstructed mesh in 8 scenes, the number of Gaussians (\#G), and training time. 
  Best results are highlighted as \colorbox{1}{ 1st }, \colorbox{2}{ 2nd }, and \colorbox{3}{ 3rd }. 
  }
  \resizebox{\linewidth}{!}{
    \tabcolsep=0.55cm
    \begin{tabular}{p{4em}|cccccccc|c|cc}
\Xhline{1pt}
\multicolumn{1}{r|}{} & 
\multicolumn{1}{p{2.5em}}{\sc{Chair}} & 
\multicolumn{1}{p{2.5em}}{\centering \sc{Drums}} & 
\multicolumn{1}{p{2.2em}}{\centering \sc{Ficus}} & 
\multicolumn{1}{p{3.4em}}{\centering \sc{Hotdog}} & 
\multicolumn{1}{p{2.2em}}{\centering \sc{Lego}} & 
\multicolumn{1}{p{4.2em}}{\centering \sc{Materials}} & 
\multicolumn{1}{p{2.2em}}{\centering \sc{Mic}} & 
\multicolumn{1}{p{2.2em}|}{\centering \sc{Ship}} & 
\multicolumn{1}{p{2.2em}|}{\centering AVG} & 
\#G &
Time \\
\hline
SpNeRF & \cellcolor{3}0.66 & 2.43 & \cellcolor{1}0.54 & \cellcolor{1}0.94 & \cellcolor{1}0.70 & 1.10 & \cellcolor{3}0.72 & 1.49 & 1.07 & - & \raisebox{0.1em}{\scalebox{0.8}{$>$}}\ 12h \\
\hline
3DGS   & 2.76 & 3.62 & 8.38 & 5.38 & 2.44 & 2.92 & 3.17 & 2.18 & 3.86 & 293k & \cellcolor{1}6.0 m \\
SuGaR  & 0.67 & \cellcolor{2}1.08 & \cellcolor{3}0.57 & 1.42 & \cellcolor{3}0.73 & 0.95 & 0.87 & \cellcolor{3}1.19 & \cellcolor{2}0.94 & -  & \raisebox{-0.2em}{\textasciitilde}\ 40 m \\
G-Surfel   & 0.75 & \cellcolor{1}0.78 & \cellcolor{2}0.56 & 1.77 & 0.93 & 1.08 & 0.99 & 2.35 & 1.15 & \cellcolor{2}128k & \cellcolor{2}7.7 m \\
2DGS   & \cellcolor{1}0.40 & 1.83 & 0.66 & 1.41 & 0.87 & \cellcolor{2}0.93 & 1.13 & \cellcolor{2}1.11 & 1.04 & \cellcolor{3}133k & 11.0 m \\
GOF    & 1.23 & \cellcolor{3}1.33 & 0.59 & \cellcolor{3}1.18 & \cellcolor{2}0.72 & \cellcolor{1}0.72 & \cellcolor{2}0.70 & 1.35 & \cellcolor{3}0.98 & 184k & 1.2h \\
\hline
Ours   & \cellcolor{2}0.47 & 1.38 & 0.69 & \cellcolor{2}1.13 & 0.81 & \cellcolor{3}0.94 & \cellcolor{1}0.61 & \cellcolor{1}0.96 & \cellcolor{1}0.87 & \cellcolor{1}69k & \cellcolor{3}10.0 m \\

\Xhline{1pt}
\end{tabular}
}%
    \label{tab:1}
\end{table*}%

\begin{table}[t]
  \centering
  \setlength{\tabcolsep}{7pt}
  \caption{Quantitative Comparison on selected scenes of Dex-NeRF dataset~\cite{ichnowskidex}. We report the Chamfer distance ($\times\ 10^{-2}$), the number of Gaussians (\#G), and training time. 
  }
  \resizebox{1\linewidth}{!}{
    \begin{tabular}{p{3.5em}|cccccc|c|cc}
    \Xhline{1pt}
          & \centering \sc{Pipe}  & \centering \sc{Mount} & \centering \sc{Up} & \centering \sc{Pawn}  & \centering \sc{Turbine} & \centering \sc{Side}    & \centering AVG  & \#G   & Time \\
    \hline
    SpNeRF & \cellcolor{3}0.82  & 1.10  & \cellcolor{2}0.83  & \cellcolor{2}0.83  & 1.27  & 1.26  & 1.02  & -     & \raisebox{0.1em}{\scalebox{0.8}{$>$}}\ 12h \\
    \hline
    3DGS  & 1.24  & 1.34  & 1.37  & 1.41  & 1.43  & \cellcolor{3}1.14  & 1.36  & 483k  & \cellcolor{3}11.4 m \\
    SuGaR & 3.36  & 1.97  & \cellcolor{3}0.84  & 2.01  & \cellcolor{2}0.79  & 3.31  & 2.05  & -     & \raisebox{-0.2em}{\textasciitilde}\ 40 m \\
    G-Surfel  & 1.41 & 1.38  & 1.29  & 1.37  & 1.30 & 1.44  & 1.37 & \cellcolor{2}47k  & 39.0 m \\
    2DGS  & \cellcolor{2}0.81  & \cellcolor{2}0.83  & 1.46  & \cellcolor{2}0.83  & 1.13  & \cellcolor{2}0.85  & \cellcolor{3}0.98  & \cellcolor{3}123k  & \cellcolor{2}9.3 m \\
    GOF   & 0.90  & \cellcolor{3}0.86  & \cellcolor{1}0.75  & \cellcolor{3}0.94  & \cellcolor{3}0.80  & 1.53  & \cellcolor{2}0.96  & 305k  & 2h \\
    \hline
    Ours  & \cellcolor{1}0.76  & \cellcolor{1}0.77  & 1.40  & \cellcolor{1}0.79  & \cellcolor{1}0.77  & \cellcolor{1}0.80  & \cellcolor{1}0.88  & \cellcolor{1}24k   & \cellcolor{1}7.5 m \\

    \Xhline{1pt}
    \end{tabular}%
  }
  \label{tab:2}
\end{table}%

\begin{table}[t]
  \centering
  \renewcommand{\arraystretch}{1.2}
  \caption{Quantitative Comparison on $15$ scenes from DTU dataset~\cite{jensen2014large}. We report the Chamfer distance, the number of Gaussians (\#G), and training time.
  }
  \resizebox{\linewidth}{!}{
  \tabcolsep=0.25cm
    \begin{tabular}{p{2em}|c|ccccc|c}
    \Xhline{1pt}
          & SpNeRF  & 3DGS & SuGaR & G-Surfel & 2DGS  & GOF   & Ours \\
    \Xhline{0.1pt}
    24    & 0.84  & 2.14  & 1.47 & 0.66  & \cellcolor{1}0.48  & \cellcolor{2}0.52  & \cellcolor{3}0.63  \\
    37    & 1.20  & 1.53  & 1.33  & \cellcolor{3}0.93 & \cellcolor{2}0.91  & \cellcolor{2}0.91  & \cellcolor{1}0.80  \\
    40    & 1.02  & 2.08  & 1.13  & 0.54 & \cellcolor{1}0.39  & \cellcolor{2}0.40  & \cellcolor{3}0.50  \\
    55    & \cellcolor{2}0.38  & 1.68  & 0.61  & 0.41 & \cellcolor{3}0.39  & \cellcolor{1}0.37  & 0.46  \\
    63    & 1.15  & 3.49  & 2.25  & \cellcolor{3}1.06 & \cellcolor{1}1.01  & 1.11  & \cellcolor{2}1.04  \\
    65    & \cellcolor{1}0.72  & 2.21  & 1.71 & 1.14  & \cellcolor{2}0.83  & \cellcolor{3}0.88  & 1.01  \\
    69    & \cellcolor{1}0.69  & 1.43  & 1.15  & 0.85 & \cellcolor{3}0.81  & \cellcolor{2}0.72  & 0.87  \\
    83    & \cellcolor{1}1.10  & 2.07  & 1.63  & 1.29 & 1.36  & \cellcolor{3}1.18  & \cellcolor{2}1.15  \\
    97    & \cellcolor{1}1.19  & 2.22  & 1.62  & 1.53 & \cellcolor{3}1.27  & \cellcolor{2}1.26  & 1.32  \\
    105   & \cellcolor{2}0.65  & 1.75  & 1.07  & 0.79 & 0.76  & \cellcolor{3}0.75  & \cellcolor{1}0.61  \\
    106   & \cellcolor{1}0.49  & 1.79  & 0.79  & 0.82 & \cellcolor{2}0.70  & \cellcolor{3}0.72  & 0.73  \\
    110   & 1.60  & 2.55  & 2.45 & 1.58  & \cellcolor{3}1.40  & \cellcolor{1}0.88  & \cellcolor{2}1.19  \\
    114   & 0.49  & 1.53  & 0.98  & \cellcolor{3}0.45 & \cellcolor{1}0.40  & 0.46  & \cellcolor{2}0.44  \\
    118   & \cellcolor{1}0.55  & 1.52   & 0.88 & \cellcolor{3}0.66 & 0.76  & 0.71  & \cellcolor{2}0.68  \\
    122   & \cellcolor{2}0.51  & 1.50  & 0.79  & 0.53 & \cellcolor{3}0.52  & 0.55  & \cellcolor{1}0.48  \\
    \Xhline{0.1pt}
    AVG   & 0.83  & 1.96  & 1.33 & 0.88  & \cellcolor{3}0.80  & \cellcolor{1}0.76  & \cellcolor{2}0.79  \\

    \hline
    \#G  & -     & 542k  & -   & \cellcolor{1}166k  & \cellcolor{3}224k  & 369k  & \cellcolor{2}171k \\
    Time  & \raisebox{0.1em}{\scalebox{0.8}{$>$}}\ 12h  & \cellcolor{2}11.2m & \raisebox{-0.2em}{\textasciitilde}\ 1h & \cellcolor{3}16.5m  & 18.8m & 1.8h  & \cellcolor{1}11.1m \\
    \Xhline{1pt}
    \end{tabular}%
  }
  \label{tab:3}
\end{table}%

\begin{figure*}[t]
    \centering
    \includegraphics[width=1\textwidth]{"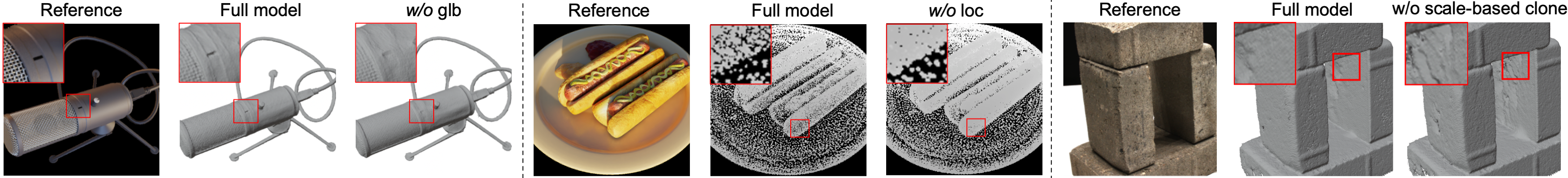"} 
    \caption{The qualitative comparison between the Spiking GS and its alternatives. 
    Columns 1-3 are the comparison of reconstructed mesh between the full model and `{\it w/o} glb'. Columns 4-6 are the comparison of Gaussians' center points projected on a specific view between the full model and `{\it w/o} loc'. Columns 7-9 are the mesh comparison between the full model and `{\it w/o} scale-based clone'.
    }
    \label{fig:abla}
\end{figure*}

\noindent\textbf{Efficiency Comparison.}
As shown in Tables~\ref{tab:1}-\ref{tab:3}, our method demonstrates a significant reduction in storage (nearly twice fewer Gaussians than 2DGS~\cite{yu2024GOF} on the NeRF-Synthetic and DTU dataset, and even five times fewer than 2DGS~\cite{yu2024GOF} on the Dex-NeRF dataset) while maintaining a competitively fast training time. 
Additionally, we also compare Spiking GS with other compression methods (C3DGS~\cite{lee2024compact} and HAC~\cite{chen2024hac}) in terms of Chamfer distance, number of Gaussians, and training time. The results are shown in \Tref{tab:4}, indicating our method balances the reconstruction accuracy, training efficiency, and storage efficiency.
In practice, many factors may affect the method's efficiency. For instance, although reducing the number of Gaussians during training can shorten training time, the additional operations (\eg, regularization computation) required to achieve this may increase time, offsetting the benefits. This explains why our method is faster than C3DGS and HAC, despite sometimes using more Gaussians. Additionally, the number of training epochs plays a critical role. For example, with a pre-trained normal estimation model, G-Surfel requires only 15000 iterations for good performance. In contrast, our method and other 3DGS approaches, lacking such priors, need more iterations. 
\subsection{Novel View Rendering}
We present the results of the novel view rendering in \Tref{tab:psnr}. Generally considering three evaluation metrics, our method shows comparable performance to G-Surfel~\cite{dai2024high}, while perform slightly worse than GOF~\cite{yu2024GOF}. We believe such a downgrading in rendered views' quality is acceptable as we use much less Gaussians to represent the scene. It also worth mentioning our method reach a superior performance in geometry reconstruction.

\subsection{Ablation Study}
In this section, we show quantitative analyses regarding the important contributions of Spiking GS (\ie, the effect of global and local FIF neurons and the scale-based clone) and used regularization losses. This results in a comparison between our method and six alternatives, as shown in Tab.\ref{tab:ablation1}. 

The results show that omitting either the global (`{\it w/o} glb') or local FIF (`{\it w/o} loc') neurons negatively impacts both the accuracy and efficiency of surface reconstruction. The two types of FIF neurons play distinct roles in our method for reducing LOPs, resulting in different effects. For instance, the global FIF neuron is primarily effective in reducing the number of LOGs to improve reconstruction quality and efficiency. Particularly, we find a method with the global FIF neuron can better capture the details on the surface, validated by a quality comparison between the full model and `{\it w/o} glb' in \fref{fig:abla}, columns 1-3. In contrast, local FIF neurons decrease LOTs to minimize the overlap of Gaussians, resulting in more separated centers of Gaussians, as evidenced by the visualized Gaussian center's distribution in \fref{fig:abla}, columns 4-6. Local FIF neurons are particularly effective in specific scenes, such as scenes with flat surfaces (the round table in \fref{fig:mainexp}, row 2) where tail overlap is predominant. Incorporating local FIF neurons will be more efficient in these cases to reduce the number of Gaussians, compared to using global FIF neurons. Removing scale-based clone (`{\it w/o} c') will cause holes-like artifacts in reconstructed surface, as shown in \fref{fig:abla}, column 9, which harms the accuracy of surface reconstruction. Moreover, the normal (`{\it w/o} $\mathcal{L}_N$'), depth distortion (`{\it w/o} $\mathcal{L}_d$'), and total variance loss (`{\it w/o} $\mathcal{L}_t$') are important empirical priors to enhance surface reconstruction quality~\cite{huang20242d, dai2024high}, which will harm the performance if we remove them. We would like to emphasize that including these priors does not diminish our contributions. While previous methods focus on refining priors to improve accuracy, they overlook the inherent deficiencies in 3DGS representations (\ie, excessive LOPs). The proposed Spiking GS improve the accuracy and efficiency of surface reconstruction in a different, novel perspective, by identifying and addressing the LOPs through spiking neurons.

\begin{table}[t]
  \centering
  \caption{PSNR, SSIM, and LPIPS scores for NeRF-Synthetic~\cite{mildenhall2021nerf} and Dex-NeRF~\cite{ichnowskidex} datasets.}
  \resizebox{\linewidth}{!}{
  \tabcolsep=0.3cm
    \begin{tabular}{l|ccc|ccc}
    \Xhline{1pt}
    Dataset & \multicolumn{3}{c|}{NeRF-Synthetic} & \multicolumn{3}{c}{Dex-NeRF} \\
    \Xhline{0.1pt}
    Metric & PSNR$\uparrow$  & SSIM$\uparrow$ & LPIPS$\downarrow$ & PSNR$\uparrow$  & SSIM$\uparrow$ & LPIPS$\downarrow$ \\
    \Xhline{0.1pt}
    3DGS  & \cellcolor{3}31.05 & \cellcolor{2}0.959 & \cellcolor{3}0.051 & 21.84 & \cellcolor{3}0.904 & 0.287  \\
    G-Surfel & \cellcolor{2}31.10  & \cellcolor{3}0.954 & \cellcolor{2}0.050  & \cellcolor{1}30.77 & \cellcolor{1}0.953 & \cellcolor{2}0.123 \\
    2DGS  & 30.28 & 0.956 & 0.054 & \cellcolor{3}22.19 & 0.886 & 0.280\\
    GOF   & \cellcolor{1}33.78 & \cellcolor{1}0.969 & \cellcolor{1}0.031 & 20.01 & \cellcolor{2}0.913 & \cellcolor{3}0.241 \\
    \Xhline{0.1pt}
    Ours  & 30.23 & \cellcolor{3}0.954 & \cellcolor{2}0.050  & \cellcolor{2}30.59 & \cellcolor{1}0.953 & \cellcolor{1}0.111 \\
    \Xhline{1pt}
    \end{tabular}%
  }
  \label{tab:psnr}%
\end{table}%

\begin{table}[t]
  \centering
  \renewcommand{\arraystretch}{1.2}
  \caption{Performance comparion between C3DGS~\cite{lee2024compact}, HAC~\cite{chen2024hac}, and Spiking GS (ours) on three datasets~\cite{ichnowskidex,jensen2014large,mildenhall2021nerf}. We report the averaged Chamfer distance (CD), the number of Gaussians (\#G), and reconstruction time. \textbf{Bold numbers} indicate the best results.}
  \resizebox{1\linewidth}{!}{
    \begin{tabular}{l|ccc|ccc|ccc}
    \Xhline{1pt}
    Dataset & \multicolumn{3}{c|}{NeRF-Synthetic} & \multicolumn{3}{c|}{Dex-NeRF} & \multicolumn{3}{c}{DTU} \\
    \Xhline{0.1pt}
    Metric & CD  & \#G & Time & CD  & \#G & Time & CD  & \#G & Time \\
    \Xhline{0.1pt}
    C3DGS & 4.59 & 118k & \textbf{10m}  & 2.41 & 62k & 12m & 2.56 & 329k & 22m \\
    HAC & 3.88 & \textbf{56k} & 13m & 4.72 & 114k & 20m & 4.33 & \textbf{61k} & 17m \\
    Ours & \textbf{0.87} & 69k & \textbf{10m} & \textbf{0.88} & \textbf{24k} & \textbf{8m}  & \textbf{0.79} & 171k & \textbf{11m} \\
    \Xhline{1pt}
    \end{tabular}%
  }
  \label{tab:4}%
\end{table}%

\begin{table}[t]
  \centering
  \renewcommand{\arraystretch}{1.1}
  \caption{Ablation study on NeRF-Synthetic dataset~\cite{mildenhall2021nerf}. We compare the proposed Spiking GS (Full) with its 6 alternatives (\ie, `w/o glb', `w/o loc', `w/o c', `w/o $\mathcal{L}_N$', `w/o $\mathcal{L}_d$', `w/o $\mathcal{L}_t$').}
  \resizebox{1\linewidth}{!}{
    \begin{tabular}{p{2em}|ccccccc}
    \Xhline{1pt}
     & Full & w/o glb  & w/o loc  & w/o c & w/o $\mathcal{L}_N$ & w/o $\mathcal{L}_d$ & w/o $\mathcal{L}_t$ \\
    \Xhline{0.1pt}
    CD$\downarrow$ & 0.87 & 0.92 & 0.88  & 0.92  & 1.94 & 0.88 & 0.88 \\
    \#G$\downarrow$ & 69k & 238k & 85k  & 64k  & 66k   & 72k & 71k \\
    \Xhline{1pt}
    \end{tabular}%
  }
  \label{tab:ablation1}%
\end{table}%

%% file: sec/6_conclusion.tex
\section{Conclusion}

We present Spiking GS, a novel method for more accurate surface reconstruction at a lower cost.
Our method reduces the number of low-opacity parts (LOPs) in the integration process for view rendering, which we have identified as one of the causes of inefficiency and reconstruction bias in Gaussian-based methods. Spiking GS incorporates global and local FIF neurons into the flattened Gaussians to address two types of LOPs: {\it low-opacity Gaussians} (LOGs) and {\it low-opacity tails} (LOTs) of Gaussians. Additionally, we implement a modified density control method and regularization losses into the training process. As a result, Spiking GS demonstrates robust performance in various datasets, achieving considerable improvements.

\noindent \textbf{Limitation.} Despite the effectiveness of the proposed Spiking GS, it still has several limitations. First, the reconstructed surface in blind spot regions exhibits strong bias, requiring additional geometric priors. Second, high-frequency details (\eg very thin structures) are challenging to reconstruct due to inherent shortcomings in mesh extraction methods. Addressing these limitations will be regarded as future work. Third, although reducing the number of LOPs generally improves the accuracy of surface reconstruction, this does not strictly happen, as discussed in \sref{sec:analyze-logs}. 
This can be further validated by LOPs count in \fref{fig:bar_chart} (c) (d) and Chamfer distance in \Tref{tab:1} (for {\sc Drums} and {\sc Mic}). That is, fewer LOPs and better performance are concurrent for most cases, a few other cases are not aligned, \ie,  our method vs. GOF,  GOF vs. G-Surfel on the {\sc Drums} object, and G-Surfel vs. GOF on the {\sc Mic} object.

\noindent \textbf{Additional Discussion.} We notice several concurrent works~\cite{fan2024trim, chen2024pgsr} that achieve impressive surface reconstruction accuracy on the DTU dataset. It is important to highlight that LOPs remain prevalent in their generated Gaussians, indicating potential for further refinement. Our method is compatible with additional losses, priors, or regularization, allowing it to improve or maintain surface reconstruction accuracy while significantly reducing the number of Gaussians. Moreover, Spiking GS demonstrates robust performance across various scenes, including challenging semi-transparent ones. In contrast, other methods may be restricted to specific scene types. Experimental comparison and more discussion can be found in the appendix.

%% file: sec/7_suppl.tex
\clearpage
\setcounter{page}{1}
\maketitlesupplementary
\setcounter{section}{6}
\setcounter{figure}{7}
\setcounter{table}{6}
\setcounter{equation}{11}

\noindent In this appendix, 
\begin{itemize}
\item we provide more training details and parameter setups in~\Sref{sec:train} (footnote 2);
\item we show more details about the regularization terms in \Sref{sec:regularization};
\item we conduct additional ablation study on the proposed spiking neurons in~\Sref{sec:more_abl}; we also test our method's compatibility to vallina 3DGS~\cite{kerbl20233d} and SOTA method~\cite{chen2024pgsr} in this section;
\item we show more discussion about the concurrent SOTA methods in~\Sref{sec:pgsr_discussion} to further clarify our contribution;
\end{itemize}
\section{Training Details and Parameter Setup}
\label{sec:train}

Following~\cite{huang20242d}, we modify the rasterization renderer to output depth and normal maps for regularization. 
For parameters setup, we follow the basic setup of learning rates in~\cite{kerbl20233d}. Particularly, to enhance training stability, we reduce the learning rates of $\bar{V}^\alpha$ and $\bar{V}^{\mathrm{p}}$ from $2e^{-4}$ to 0 during first 300 iterations of every 3000 iterations within the initial 15000 iterations. We empirically using three different $\bar{V}^\theta$ for all three datasets, $\bar{V}^\theta = 0.02$ for scenes in NeRF-Synthetic, $\bar{V}^\theta = 1$ for scenes in Dex-NeRF, and $\bar{V}^\theta = 0.01$ for scenes in DTU. 
The settings are based on our observations: 1) for scenes in Dex-NeRF, we observe smoother surfaces, so a higher parameter is preferred. 2) for scenes in NeRF-Synthetic and DTU, we observe more intricate surfaces, so a lower parameter is preferred. 
For mesh extraction during evaluation, we use truncated signed distance fusion (TSDF) to extract meshes from depth maps~\cite{huang20242d}. In our implementation, we set the voxel size to $0.004$ and the truncation threshold to $0.02$. All of our experiments are conducted on a single 24GB NVIDIA RTX3090 GPU.

\section{Details of Regularization Terms}
\label{sec:regularization}
\textbf{Depth distortion loss.}
Optimizing solely on $\mathcal{L}_{c}$~\cite{kerbl20233d} can result in noisy surfaces, so we follow~\cite{huang20242d} to introduce depth distortion loss. Depth distortion loss $\mathcal{L}_d$ reduces the depth disparity along the ray, concentrating the Gaussian splats to be closer to each other:
\begin{equation}
    \label{eq:13}
    \mathcal{L}_d=\sum_{i,j}\lambda^d_i\lambda^d_j|t_i-t_j|,
\end{equation}
where $\lambda^d_i=\alpha^\prime_i \mathcal{G}^{\prime}_i(\mathbf{x}) \prod_{j=1}^{i-1}\left(1-\alpha_j^\prime \mathcal{G}_j^{\prime}(\mathbf{x})\right)$ is the blending weight of $i$-th Gaussian and $t_i$ is its depth. While the depth of Gaussian' center $\mathbf{p}$ contain biases~\cite{dai2024high}, we follow~\cite{dai2024high} to use the depth at the ray-Gaussian intersection for $\mathcal{L}_d$. 

\noindent \textbf{Normal loss.} Normal consistency loss $\mathcal{L}_n$~\cite{huang20242d} helps the Gaussians to align with the actual surfaces by ensuring consistency between the Gaussians' normal and the surface normal:
\begin{equation}
    \label{eq:14}
    \mathcal{L}_N=\sum_{i}\lambda^N_i(1-\mathbf{n}_i^\top \mathbf{N}),
\end{equation}
where $\mathbf{n}_i$ denotes the normal of Gaussian and $\mathbf{N}$ is the surface normal estimated with the gradients of the depth maps~\cite{huang20242d}. To smooth the estimated surface normal, we further apply a bilateral filter~\cite{elad2002origin} on the depth maps.

\noindent\textbf{Total variance loss.} Following ~\cite{karnieli2022deepshadow,turkulainen2024dn}, we apply the edge-aware total variance loss on depth maps to smooth the surface representation:
\begin{equation}
    \label{eq:18}
    \mathcal{L}_t=\sum_{ij} |\partial_x\hat{d}_{ij}|e^{-\|\partial_x\Bar{I}_{ij}\|} + |\partial_y\hat{d}_{ij}|e^{-\|\partial_y\Bar{I}_{ij}\|},
\end{equation}
where $\partial_x$ and $\partial_y$ are the gradients in the horizontal and vertical directions, $\hat{d}_{ij}$ is the estimated depth at pixel $(u_i,v_j)$ on depth maps, and $\Bar{I}$ is the average color of ground truth images. This regularization term improves smoothness of depth maps while offers a faster convergence. 

\section{Additional Ablation Studies}
\label{sec:more_abl}
\begin{table}[t]
  \centering
  \renewcommand{\arraystretch}{1.1}
  \caption{Ablation study on NeRF-Synthetic dataset~\cite{mildenhall2021nerf}. We compare the proposed Spiking GS (Full) with its 4 alternatives (\ie, `w/o $\mathcal{L}_\alpha$', `w/o $\mathcal{L}_\mathrm{p}$', `w/o $\mathcal{L}_\mathrm{s}$', `w/ $\mathrm{GP}$') and 3DGS with its 2 alternatives (\ie, `w/ glb', `w/ loc').}
  \resizebox{1\linewidth}{!}{
    \begin{tabular}{p{2em}|ccccc|cccc}
    \Xhline{1pt}
     & Full & w/o $\mathcal{L}_\alpha$  & w/o $\mathcal{L}_p$  & w/o $\mathcal{L}_s$ & w/ GP & 3DGS & w/ glb & w/ loc \\
    \Xhline{0.1pt}
    CD$\downarrow$ & 0.87 & 0.91  & 0.89  & 1.03 & 1.07 & 3.86 & 3.72 & 2.88  \\
    \#G$\downarrow$ & 69k & 245k  & 74k  & 55k   & 68k  & 293k & 55k & 211k \\
    \Xhline{1pt}
    \end{tabular}%
  }
  \label{tab:ablation2}%
\end{table}%

\noindent\textbf{The effectiveness of proposed spiking neuron} is further validated by a comparison with the vanilla 3DGS and two alternatives, `w/ glb' and `w/ loc' in \Tref{tab:ablation2}, where only the spiking neurons are additionally implemented and the original 3DGS setup is retained. We observe a considerable improvement in both accuracy and memory efficiency.

\begin{figure}[t]
    \centering
    \includegraphics[width=0.48\textwidth]{"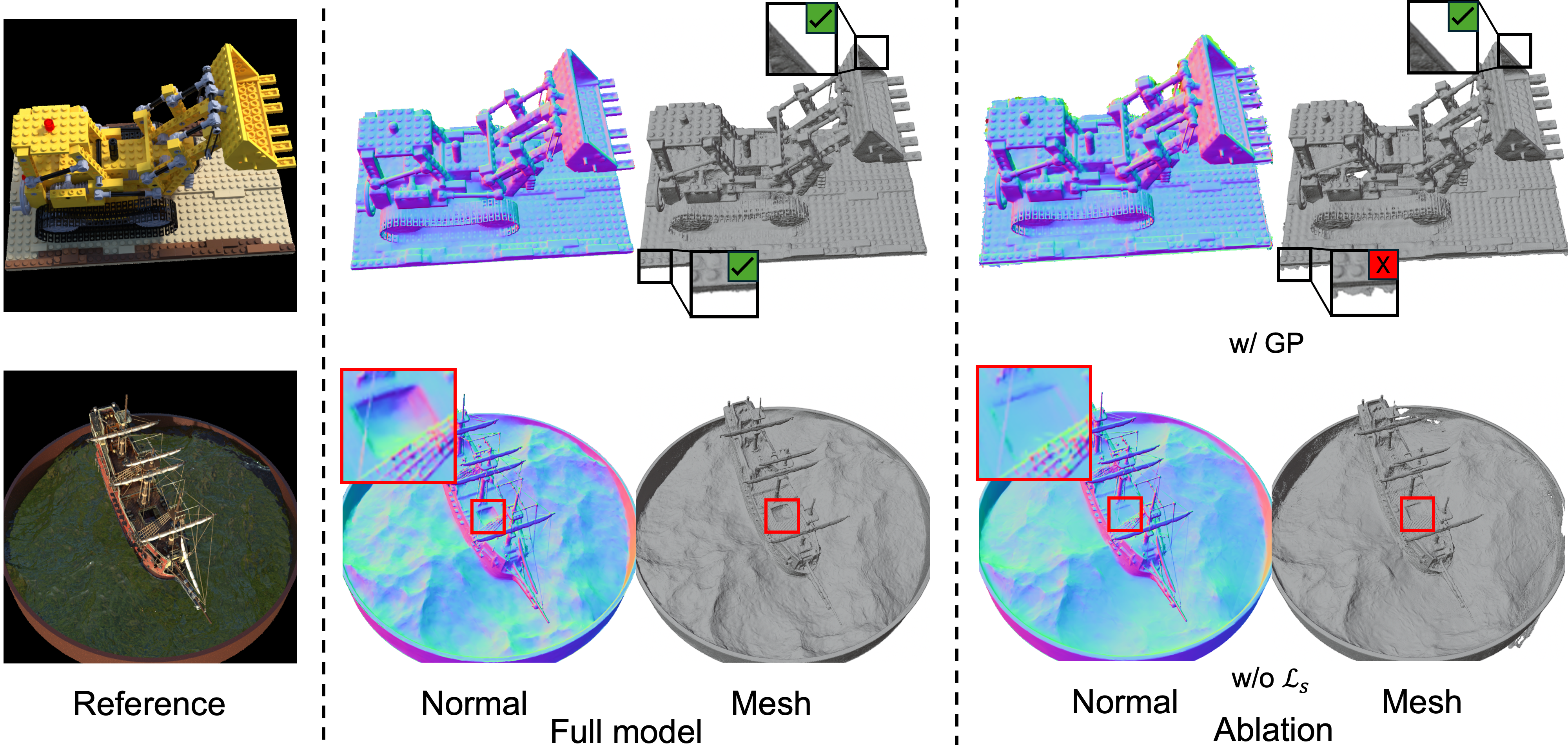"} 
    \caption{The qualitative comparison between the Spiking GS and its alternatives. From top to down: ablation on the necessity of a local FIF neuron on each Gaussian representation function and ablation on scale loss $\mathcal{L}_\mathrm{s}$.
    }
    \label{fig:abla2}
\end{figure}

\noindent\textbf{Necessity of a local FIF neuron on each Gaussian representation function} is validated by comparison with the alternative, `w/ $\mathrm{GP}$'. Specifically, `w/ $\mathrm{GP}$' uses a global threshold shared by all Gaussians.  As shown in~\Tref{tab:ablation2}, a shared threshold will dramatically downgrade the accuracy of the reconstructed surface. Through a qualitative comparison (\fref{fig:abla2}), we identify that the downgrading in accuracy is caused by unnecessary extension of surface (\eg, the base board of {\sc Lego}). The analysis above indicates the necessity of an isolate FIF neuron on each Gaussian representation function to fit the geometry at different locations.

\noindent\textbf{Effectiveness of proposed loss} is assessed by comparison between Spiking GS and its three alternatives (\ie, `w/o $\mathcal{L}_\alpha$', `w/o $\mathcal{L}_\mathrm{p}$', and `w/o $\mathcal{L}_\mathrm{s}$'). As can be seen on \Tref{tab:ablation2}, both the number of Gaussians and Chamfer distance increase without $\mathcal{L}_\alpha$ and $\mathcal{L}_\mathrm{p}$, proving the loss on $\bar{V}^\alpha$ and $\bar{V}^\mathrm{p}$ boosting the effect of FIF neurons.  Additionally, a quantitative (\Tref{tab:ablation2}) and qualitative (\fref{fig:abla2}) comparison between the full model and `w/o $\mathcal{L}_\mathrm{s}$' demonstrate the effect of $\mathcal{L}_\mathrm{s}$ in improving surface's details.

\noindent\textbf{Compatibility of proposed method} is evaluated by comparison between several baseline methods and their alternatives. Specifically, we select both representative work (\ie, 3DGS~\cite{kerbl20233d}) and concurrent SOTA work (\ie, PGSR~\cite{chen2024pgsr}) as the baseline, incorporating our two types of FIF spiking neurons into their pipeline (denoted as S-3DGS and S-PGSR, respectively). The hyperparameter of spiking neurons is set to the values used in our experiments, while other parameters follow the default settings of the original methods. We skip to report the results on challenging Dex-NeRF~\cite{ichnowskidex} dataset as baseline methods perform poorly in reconstructing semi-transparent objects. As shown in \Tref{tab:pgsr}, the improvement on the accuracy and efficiency are highly competitive. The decreasing number of Gaussians highlights the prevalence of LOPs in previous methods. The overall performance demonstrates the generality and compatibility of our FIF spiking neurons across Gaussian-based methods. However, as discussed in the main paper, the improvement diminishes on the DTU dataset due to the concentrated views from the front of the object and the lack of coverage at the back.

\section{Additional Discussion}
\label{sec:pgsr_discussion}
Although some concurrent works~\cite{fan2024trim, chen2024pgsr} reconstruct accurate geometry on the DTU dataset~\cite{jensen2014large}, we find that these methods still suffer from issues caused by excessive LOPs and poor reconstruction results on some challenging scenes (\eg, semi-transparent objects from the Dex-NeRF dataset), where both methods fail to generate reasonable results.

\noindent \textbf{Discussion about TrimGS~\cite{fan2024trim}}. TrimGS~\cite{fan2024trim} introduced a novel density control strategy to trim inaccurate Gaussians based on a pre-trained Gaussian model. However, it tends to overly split and generate numerous Gaussians if the pre-trained model contains excessively large LOPs. According to our experiment, the number of Gaussians of the trimmed 2DGS~\cite{fan2023lightgaussian,huang20242d} could exceed ten million in NeRF-Synthetic~\cite{mildenhall2021nerf} and Dex-NeRF~\cite{ichnowskidex} datasets, which severely undermine training efficiency and consume a significant amount of VRAMs.

\noindent \textbf{Discussion about PGSR~\cite{chen2024pgsr}}. PGSR~\cite{chen2024pgsr} utilized a multiview geometry consistency prior constraint to regularize the reconstructed surface, exhibiting strong performance in smooth surface reconstruction. Nevertheless, they overlooked the prevalence of LOPs and the issues associated with. Our method can be implemented into their pipeline. Specifically, we integrate our FIF spiking neurons into their method. Through a quantitative comparison among our method, the original PGSR, and PGSR with spiking neurons (S-PGSR) on the NeRF-Synthetic and DTU datasets in \Tref{tab:pgsr}, we observe an improvement in reconstruction accuracy and efficiency in surface reconstruction, further validating the need to reduce the number of LOPs.

\begin{table}[t]
  \centering
  \renewcommand{\arraystretch}{1.1}
  \caption{Additional quantitative Comparison on NeRF-Synthetic~\cite{mildenhall2021nerf} and DTU~\cite{jensen2014large} datasets between Spiking GS, 3DGS~\cite{kerbl20233d}, PGSR~\cite{chen2024pgsr}, and their alternatives (S-3DGS and S-PGSR, integrated with our method). We report the averaged Chamfer distance (CD), the number of Gaussians (\#G), and reconstruction time. \textbf{Bold numbers} indicate the best results between the baseline and corresponding alternative.}
  \resizebox{0.9\linewidth}{!}{
    \begin{tabular}{l|ccc|ccc}
    \Xhline{1pt}
    Dataset & \multicolumn{3}{c|}{NeRF-Synthetic} & \multicolumn{3}{c}{DTU} \\
    \Xhline{0.1pt}
    Metric & CD$\downarrow$  & \#G$\downarrow$ & Time$\downarrow$ & CD$\downarrow$ & \#G$\downarrow$ & Time$\downarrow$ \\
    \Xhline{0.1pt}
    Ours &  0.87 & 69k & 10.0m & 0.79 & 171k & 11.1m  \\
    \Xhline{0.1pt}
    3DGS & 3.86 & 293k & 6.0m & 1.96 & 542k & 11.2m  \\
    S-3DGS & \textbf{2.80} & \textbf{57k} & \textbf{5.4m} & \textbf{1.83} & \textbf{186k} & \textbf{9.2m}  \\
    \Xhline{0.1pt}
    PGSR & 0.84 & 205k & 24.3m & \textbf{0.55} & 165k & 27.6m  \\
    S-PGSR & \textbf{0.81} & \textbf{77k} & \textbf{23.5m} & 0.56 & \textbf{88k} & \textbf{24.1m}  \\
    \Xhline{1pt}
    \end{tabular}%
  }
  \label{tab:pgsr}%
\end{table}%

\begin{table}[t]
  \centering
  \caption{PSNR, SSIM, and LPIPS scores for DTU~\cite{jensen2014large} dataset in common (12.5\% of images for testing) and sparse (50\% for testing) settings. Best results are highlighted as \colorbox{1}{ 1st }, \colorbox{2}{ 2nd }, and \colorbox{3}{ 3rd }.}
  \resizebox{\linewidth}{!}{
  \tabcolsep=0.3cm
    \begin{tabular}{l|ccc|ccc}
    \Xhline{1pt}
    Dataset & \multicolumn{3}{c|}{12.5\%} & \multicolumn{3}{c}{50\%} \\
    \Xhline{0.1pt}
    Metric & PSNR$\uparrow$  & SSIM$\uparrow$ & LPIPS$\downarrow$ & PSNR$\uparrow$  & SSIM$\uparrow$ & LPIPS$\downarrow$ \\
    \Xhline{0.1pt}
    3DGS  & \cellcolor{2}32.99 & \cellcolor{2}0.957 & \cellcolor{3}0.059 & \cellcolor{3}32.25 & 0.948 & \cellcolor{2}0.060 \\
    G-Surfel & 32.51 & 0.942 & 0.060 & 31.70 & 0.936 & 0.063 \\
    2DGS  & 32.39 & 0.955 & \cellcolor{2}0.058 & 31.99 & \cellcolor{3}0.949 & \cellcolor{3}0.061 \\
    GOF   & \cellcolor{1}33.01 & \cellcolor{1}0.960 & \cellcolor{1}0.050 & \cellcolor{1}32.36 & \cellcolor{1}0.953 & \cellcolor{1}0.055 \\
    \Xhline{0.1pt}
    Ours  & \cellcolor{3}32.59 & \cellcolor{3}0.956 & \cellcolor{3}0.059 & \cellcolor{2}32.31 & \cellcolor{2}0.951 & 0.062 \\
    \Xhline{1pt}
    \end{tabular}%
  }
  \label{tab:psnr2}%
\end{table}%
\begin{figure*}[t]
    \centering
    \includegraphics[width=0.95\textwidth]{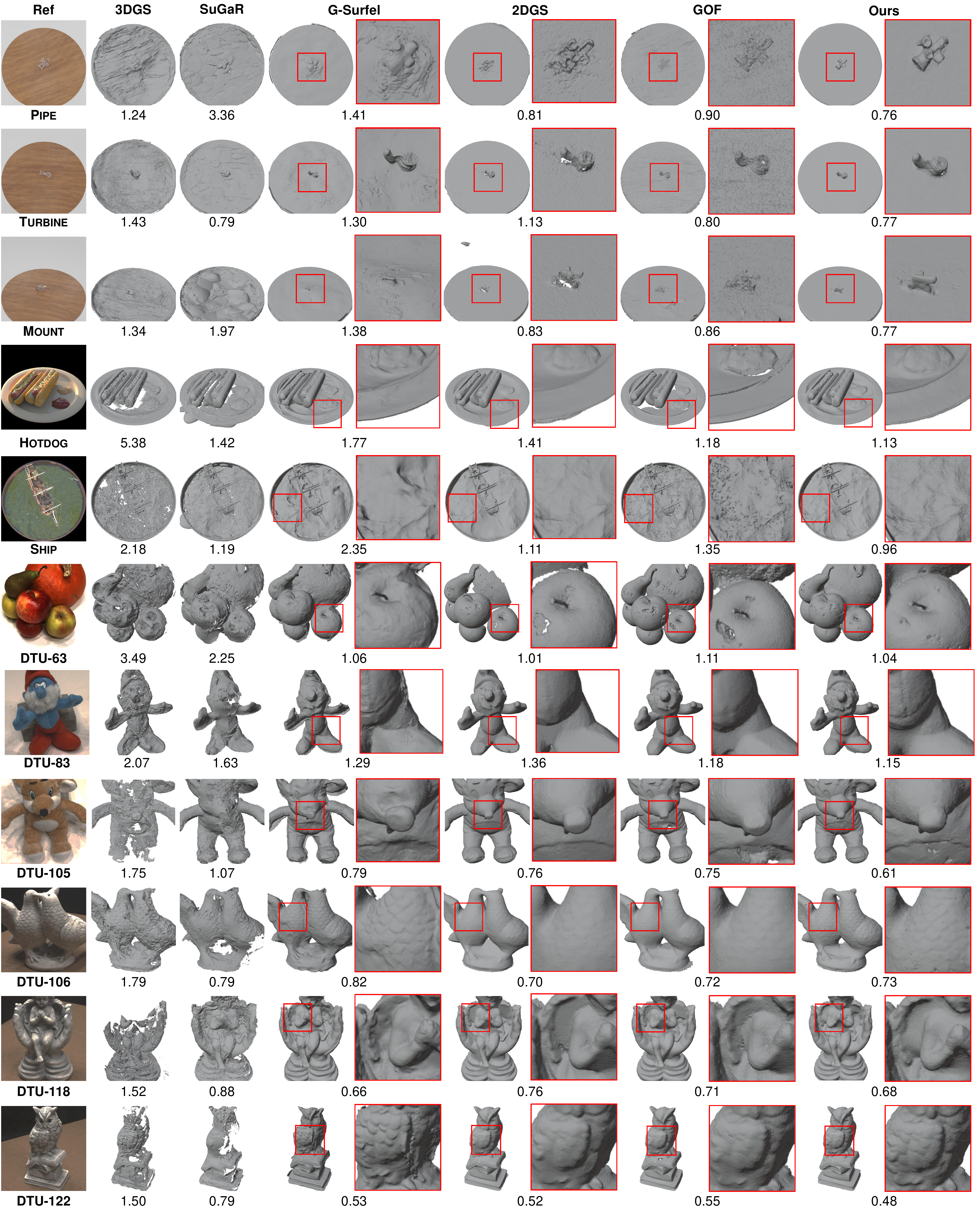}
    \caption{\textbf{Additional qualitative comparisons of surface reconstruction} performed on Dex-NeRF~\cite{ichnowskidex} (row 1 to 3), NeRF-Synthetic~\cite{mildenhall2021nerf} (row 4 and 5), and DTU~\cite{jensen2014large} (row 6 to 11) datasets. We show the Chamfer distance in the bottom of the image. 
}
\label{fig:supp_mainexp}
\end{figure*}

\section{Additional Results}

We present additional rendering quality comparisons result on DTU under two conditions: a common setting with 12.5\% of images reserved for testing, and a sparse setting with 50\% of images reserved for testing. As shown in \Tref{tab:psnr2}, our method exhibits better rendering generality compared to both 3DGS and 2DGS. In \Fref{fig:supp_mainexp}, we shown additional qualitative comparisons result on Dex-NeRF dataset~\cite{ichnowskidex}, NeRF-Synthetic dataset~\cite{mildenhall2021nerf}, and DTU dataset~\cite{jensen2014large}.